\documentclass{article}

\usepackage{arxiv}
\usepackage{authblk}
\usepackage[utf8]{inputenc} 
\usepackage[T1]{fontenc}    
\usepackage{hyperref}       
\usepackage{url}            
\usepackage{booktabs}       
\usepackage{amsfonts}       
\usepackage{nicefrac}       
\usepackage{microtype}      
\usepackage{lipsum}		
\usepackage{graphicx}
\usepackage[numbers]{natbib}
\usepackage{doi}
\usepackage{bookmark}

\title{SyncFed: Time-Aware Federated Learning through Explicit Timestamping and Synchronization}

\author[1]{\textit{Baran Can Gül}}
\author[1,2]{\textit{Stefanos Tziampazis}}
\author[1]{Nasser Jazdi}
\author[1]{Michael Weyrich}

\affil[1]{\footnotesize Institute of Industrial Automation and Software Engineering, University of Stuttgart, Germany}
\affil[ ]{{\footnotesize
  \texttt{\{baran‑can.guel, nasser.jazdi, michael.weyrich\}@ias.uni-stuttgart.de}
}}
\affil[2]{\footnotesize Graduate School of Excellence advanced Manufacturing Engineering, University of Stuttgart, Germany}

\affil[ ]{{\footnotesize
  \texttt{stefanos.tziampazis@gsame.uni-stuttgart.de \vspace{-0.5em}}
}}
\affil[ ]{}
\affil[ ]{\footnotesize \textit{The first two authors contributed equally to this work.\vspace{-2.5em}}}

\date{}

\renewcommand{\headeright}{}
\renewcommand{\undertitle}{}
\renewcommand{\shorttitle}{\textit{Preprint version. Accepted for publication at IEEE ETFA 2025.}}

\hypersetup{
pdftitle={A template for the arxiv style},
pdfsubject={q-bio.NC, q-bio.QM},
pdfauthor={David S.~Hippocampus, Elias D.~Striatum},
pdfkeywords={First keyword, Second keyword, More},
}

\begin{document}

\begingroup

\vspace*{-2cm}  
\begin{center}
    {\itshape Preprint version. Accepted for publication at IEEE ETFA 2025.}
\end{center}
\maketitle
\endgroup

\begin{abstract}
As Federated Learning (FL) expands to larger and more distributed environments, consistency in training is challenged by network-induced delays, clock unsynchronicity, and variability in client updates. This combination of factors may contribute to misaligned contributions that undermine model reliability and convergence. Existing methods like staleness-aware aggregation and model versioning address lagging updates heuristically, yet lack mechanisms to quantify staleness, especially in latency-sensitive and cross-regional deployments. In light of these considerations, we introduce \emph{SyncFed}, a time-aware FL framework that employs explicit synchronization and timestamping to establish a common temporal reference across the system. Staleness is quantified numerically based on exchanged timestamps under the Network Time Protocol (NTP), enabling the server to reason about the relative freshness of client updates and apply temporally informed weighting during aggregation.
Our empirical evaluation on a geographically distributed testbed shows that, under \emph{SyncFed}, the global model evolves within a stable temporal context, resulting in improved accuracy and information freshness compared to round-based baselines devoid of temporal semantics.
\end{abstract}


\section{Introduction}
Federated Learning (FL) has become a foundational approach for enabling collaborative model development while preserving data locality. It has seen broad adoption across domains such as the \textit{Internet of Things} (IoT)~\cite{intro1}, \textit{smart grids}~\cite{intro2}, \textit{vehicular networks}~\cite{intro3, B1} , and \textit{healthcare}~\cite{intro4}. These settings share a reliance on \emph{geographically distributed} nodes that operate under regulatory and infrastructural constraints; model training occurs locally, while a central server aggregates the resulting model updates, rather than the raw data itself.

The very decentralization that underpins FL also introduces structural limitations. Communication bottlenecks, network heterogeneity, and intermittent connectivity are known sources of instability, particularly in cross-regional deployments with lossy links. These conditions frequently lead to inconsistencies: model updates arriving at the central server are computed using \textit{stale} local parameters, a phenomenon referred to as \textit{staleness}. In such cases, updates reflect outdated model states that lag multiple steps behind the current global model, thereby impairing convergence. \textit{Staleness} has been examined in FL literature through coordination strategies that are typically classified by their temporal structure: \textit{synchronous}, \textit{asynchronous}, and \textit{semi-synchronous}~\cite{intro5}. \textit{Synchronous} FL waits for all, or a designated subset of clients, before aggregation, offering consistency but incurring latency and sensitivity to stragglers. \textit{Asynchronous} schemes process updates as they arrive, improving responsiveness while intensifying staleness and inconsistency. \textit{Semi-synchronous} methods seek a compromise by aggregating updates from a fixed number of clients or within bounded time windows.

Most of these techniques address staleness heuristically: updates are down-weighted or delayed based on estimated staleness, typically approximated via round counts. This reveals an intrinsic limitation of many FL workflows: \textit{the absence of a unified reference for time coordination}, rooted in an inherently untimed view of staleness. Staleness cannot be quantified consistently and without explicit temporal alignment, FL deployments must navigate a trade-off between waiting for slower clients or forging ahead with stale updates, a tension that becomes acute in geographically distributed and latency-sensitive environments.


Temporal challenges in such cross-regional deployments are not unique to FL; in the broader field of distributed systems, time coordination has long been a topic of sustained research, giving rise to a range of time-synchronization techniques. Mechanisms such as the Network Time Protocol (NTP)~\cite{NTP}, the Precision Time Protocol (PTP)~\cite{PTP}, and GPS-based timekeeping are commonly used to align clocks across nodes in order to support event sequencing and timely system behavior. Building on these established strategies, we introduce \textbf{SyncFed}, a time-aware FL framework that bridges federated learning with distributed systems synchronization.  Specifically, \textbf{SyncFed} incorporates explicit timestamping into the FL workflow: each update is generated, labeled, and aggregated according to a \textit{shared clock reference} managed via NTP. By quantifying staleness numerically, the global server applies \textit{temporally informed weighting}, to reduce the impact of out-of-date contributions, ensuring that fresher updates drive the model more decisively. We envisage that this time-aware coordination approach can provide finer-grained control over staleness than heuristic round-based methods and offer improved scalability in latency-sensitive environments. 


The remainder of this paper is organized as follows. Section~\ref{related} reviews existing works on federated learning paradigms, staleness mitigation strategies, and synchronization principles. Section~\ref{framework} details the \textbf{SyncFed} architecture with a focus on the explicit timestamping and synchronization components. Section~\ref{implementation} reports on the empirical evaluation, while Section~\ref{discussion} discusses implications and future considerations. Finally, Section~\ref{conclusion} concludes the paper by summarizing key contributions.


\section{Related Work}
\label{related}
In this section, we review prior work on temporal aspects of FL and related themes in distributed systems. We begin with the notion of \textit{staleness} and coordination strategies designed to mitigate it. We then discuss concepts such as \textit{timeliness }and \textit{Age of Information} (AoI), followed by an overview of synchronization mechanisms in distributed computing. Finally, we outline existing approaches integrating \textit{time awareness} or \textit{delay sensitivity} in FL frameworks.

\subsection{Staleness in Federated Learning}
Staleness in FL deployments arises naturally from the decentralized nature of model training: each client performs local training independently and transmits updates to a central server for aggregation. Factors such as communication latency, variable client availability, and heterogeneity in computational capacity may cause updates to arrive at different times, resulting in gradients computed on outdated model parameters~\cite{r1}. These so-called \textit{stale updates} can distort learning dynamics, hinder convergence, and lead to suboptimal accuracy, particularly in large-scale or cross-regional deployments with geographically dispersed clients~\cite{r2}. FL literature has proposed various approaches to address staleness, each with its own trade-offs: (\textit{a}) \textit{\textbf{Synchronous FL}}: The server waits for all or a designated subset of clients before aggregating updates~\cite{r3,r4,r5}. While this ensures version consistency, it is highly sensitive to stragglers and latency. (\textit{b}) \textit{\textbf{Asynchronous FL}}: Updates are applied as they arrive, without waiting for slower clients~\cite{r6,r7}. This improves responsiveness, but makes stale updates commonplace and, without mitigation, can lead the server to incorporate updates based on significantly outdated model states. (\textit{c}) \textit{\textbf{Semi-Synchronous FL}}: These techniques aim to strike a balance between strict synchrony and full asynchrony by aggregating updates from a subset of clients or within a fixed time window~\cite{r8,r9,r10}, trading faster turnaround for partial control over staleness. Although these three paradigms differ in how they manage waiting and update timing, most rely on coarse heuristics, such as round counters, to detect staleness. This reliance highlights a broader limitation: the lack of a more precise mechanism for assessing the actual age or \textit{freshness} of client contributions.

\subsection{Timeliness and Age of Information in Federated Learning}

In the broader field of timed systems, the notion of time-sensitivity in information exchange is captured by different metrics. Two closely related concepts are \textit{\textbf{timeliness}}, which assesses how quickly new data is propagated, and \textit{\textbf{Age~of~Information}} (AoI), which gauges how \textit{old} or \textit{recent} the information at a receiver is relative to when it was generated.  Building on these notions, the authors of~\cite{r12} introduced an AoI-aware federated framework that incentivizes clients to utilize fresh local datasets, formulating client selection as a restless multi-armed bandit problem and employing Whittle’s Index to balance dataset currency with a constrained budget. Authors in \cite{r12} proposed an age-weighted Federated Stochastic Gradient Descent (FedSGD) algorithm to mitigate weight divergence in non-independent and identically distributed (non-IID) settings by scaling gradients according to each device’s participation history, coupled with a resource allocation strategy that reduces energy consumption and improves device participation. Another reference \cite{r13} presented FedAoI, a client selection policy that minimizes Peak Age of Information (Peak-AoI) to balance fairness with efficiency in heterogeneous wireless settings. While AoI-based approaches have shown promise in improving client selection and participation fairness, they do not address the underlying temporal inconsistency in update integration, as they typically rely on local or heuristic indicators of staleness and lack a globally shared notion of time.

\subsection{Synchronization in Distributed Systems and Federated Learning}
The design of distributed systems has long relied on time synchronization to consistently timestamp events and maintain a coherent global order, typically utilizing established protocols like the Network Time Protocol (NTP)~\cite{NTP} and the Precision Time Protocol (PTP)~\cite{PTP}. Standards such as Time-Sensitive Networking (TSN)~\cite{TSN} further incorporate synchronization (e.g., gPTP) alongside other protocols to guarantee deterministic communication and timing. However, the adaptation of these concepts to FL remains an emerging topic within the field. 
Some initial work has explored synchronization in FL, including clock offset inference and time-aware aggregation. For instance, works in~\cite{lyu2020time} and~\cite{chen2021timestamping} highlight the importance of timestamping for ensuring consistent update handling, while research in~\cite{zhu2022timeaware} uses Age of Information (AoI) to prioritize updates based on their freshness. In industrial settings, a fault-tolerant FL framework over TSN is proposed in~\cite{r14}, leveraging Software-Defined Networking for recovery to minimize joint failure probabilities. Similarly, a three-layer deterministic FL architecture is introduced in~\cite{r15}, utilizing TSN scheduling to achieve ultra-reliable, low-latency model updates under stringent Quality of Service (QoS) requirements.
While effective in controlled environments, such TSN-based methods face limitations in cross-regional deployments due to hardware and distance constraints. 
NTP-based synchronization has also been considered. 
For example, a learning-based synchronization technique for hierarchical FL aims to optimize aggregation timing by implicitly using NTP-like principles~\cite{r16}. Another investigation addresses synchronization in energy-limited hierarchical FL settings, while conceding that explicit time coordination remains challenging~\cite{r17}. 
Despite these efforts, concrete implementations of NTP in FL remain limited; most refer to synchronization conceptually without providing concrete implementation details.

\subsection{Time-Aware or Delay-Sensitive Federated Learning}
While fully time-synchronized federated learning remains an emerging topic, a few studies have introduced limited forms of time awareness. For instance, staleness-aware aggregation techniques adjust the weighting of updates based on their age, employing methods such as exponentially decaying factors or polynomial penalties~\cite{r18}. Other frameworks estimate network conditions, like round-trip time, to implement scheduling policies that prioritize faster responders, thereby mitigating straggler effects~\cite{r19}. However, these methods often lack an explicit, system-wide clock reference to which staleness can be precisely mapped. A related line of inquiry focuses on delay-sensitive FL, where the training process is tailored to meet real-time constraints or critical latencies~\cite{r20}; yet even here, parameters such as deadlines or maximum lag are imposed without establishing an explicit system-wide synchronization among clients. 

\vspace{1em}
To date, an effort to amalgamate these aspects into a cohesive, time-aware federated learning framework that incorporates explicit synchronization, a unified temporal reference, and quantitative measures of update freshness has, to the best of our knowledge, not yet been undertaken.

\section{SyncFed: A Framework for Freshness-weighted Synchronized Time-aware Federated Learning}
\label{framework}



\begin{figure}[t]
\centering
\includegraphics[width=3.5in]{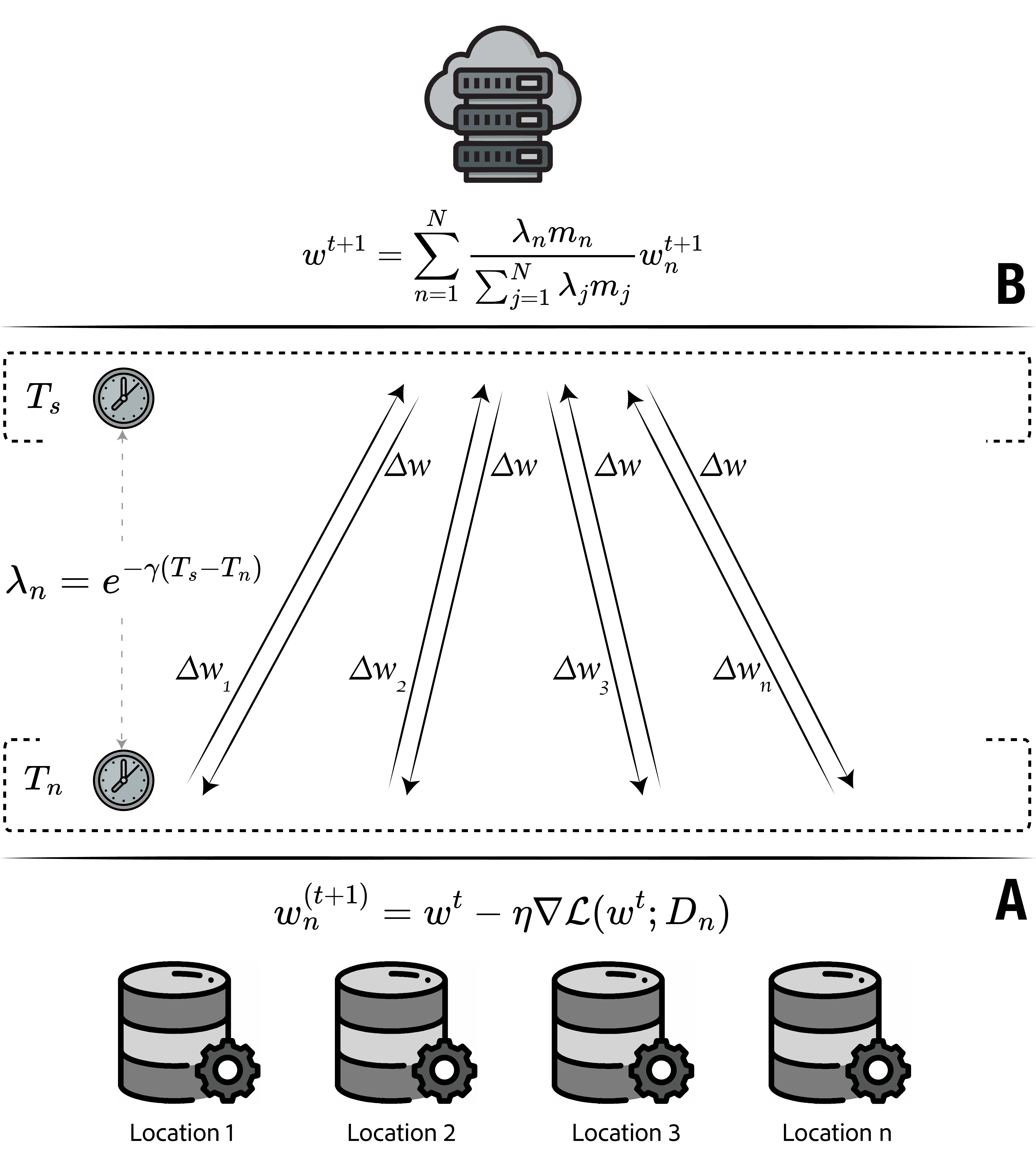}
\caption{Overview of the \textbf{SyncFed} Framework:
(A) Distributed clients perform local training and timestamp updates.
(B) The server computes staleness from synchronized timestamps and applies freshness-weighted aggregation.}
\label{Figure_section_3}
\end{figure}

In this section, we detail \textbf{SyncFed} and provide a formulation of its key mechanisms, with an emphasis on how synchronization and freshness are incorporated into the training process. An overview of the framework is shown in Fig.~\ref{Figure_section_3}, with its main components detailed in the subsections that follow. Section \ref{local_training} describes the local training procedure and timestamping at each client. Section \ref{update_to_server} explains how update freshness is measured and communicated and introduces the aggregation method, which weights updates based on both recency and data volume.

\subsection{Local Training at Each Node}
\label{local_training}
Each client connected to the network performs local model training using its private dataset \( D_n \) and the most recent global model \( w^t \) received from the server. After performing stochastic gradient descent (SGD) on the local data, the node updates its model and sends the update, along with a corresponding timestamp, back to the server. Specifically, for node \( n \), the update is computed as follows:

\begin{equation}
    w_n^{(t+1)} = w^t - \eta \nabla \mathcal{L}(w^t; D_n)
\end{equation}
where \( w_n^{(t+1)} \) is the locally updated model at node \( n \), \( w^t \) is the received global model at round \( t \), \( \eta \) is the learning rate, and \( \nabla \mathcal{L}(w^t; D_n) \) is the gradient of the loss function \( \mathcal{L} \) computed on the local dataset \( D_n \).

\subsection{Sending Update to the Server}
\label{update_to_server}
After completing local training, each client sends the model update \( \Delta w_n = w_n^{(t+1)} - w^t \) along with a timestamp \( T_n \), indicating the time when the local update was computed. The server uses this timestamp to calculate the \textit{freshness weight} \( \lambda_n \), which quantifies the recency of each update. The freshness weight is computed as:

\begin{equation}
    \lambda_n = e^{-\gamma (T_s - T_n)}
\end{equation}
where \( T_s \) is the current time at the server (synchronized using NTP), and \( T_n \) is the timestamp of the client's model update. The decay factor \( \gamma \) controls the rate at which the weight decreases for stale updates: smaller values of \( \gamma \) allow older updates to have a higher impact, while larger values more aggressively reduce the impact of outdated contributions.

In traditional Federated Averaging (FedAvg), the global model update is computed as a weighted average of the local model updates, where the weights are typically proportional to each client's dataset size \( m_n \).

\begin{equation}
    w^{t+1} = \sum_{n=1}^{N} \frac{m_n}{M} w_n^{t+1}
\end{equation}
where \( M = \sum_{n=1}^{N} m_n \) is the total number of data points across all clients. In this scheme, all local updates are treated equally, with no regard for the freshness of the updates.

In contrast, \textit{SyncFed} extends the FedAvg algorithm by incorporating the freshness weight \( \lambda_n \) into the aggregation process. This modification prioritizes updates from clients with more recent data during model aggregation. The global model update in SyncFed is computed as:

\begin{equation}
    w^{t+1} = \sum_{n=1}^{N} \frac{\lambda_n m_n}{\sum_{j=1}^{N} \lambda_j m_j} w_n^{t+1}
\end{equation}
where the numerator \( \lambda_n m_n \) combines both the dataset size and freshness for each client. The denominator serves to normalize the weights across all clients. In this approach, more recent updates (higher \( \lambda_n \)) exert greater influence on the global model, while outdated updates (lower \( \lambda_n \)) have a reduced contribution, thereby limiting the effect of stale information. This approach is especially effective in environments characterized by intermittent connectivity or variable availability, such as vehicular networks, where timely updates are crucial.

\vspace{1em}
The contributions of SyncFed can be summarized as follows:

\begin{itemize}
    \item \textbf{Freshness Weighting}: Updates are weighted based on their recency, so fresher updates have more influence on the global model.
    \item \textbf{Explicit time Synchronization}: NTP ensures a shared clock across clients, allowing update staleness to be measured instead of estimated heuristically.
    \item \textbf{Time-aware Aggregation}: NTP ensures a shared clock across clients, allowing update staleness to be measured accurately instead of estimated heuristically.
\end{itemize}

\section{Implementation and Experimental Setup}
\label{implementation}

To evaluate the effectiveness of our proposed \textbf{SyncFed} approach in handling staleness during the learning process, we designed a time-aware experimental setup simulating real-world vehicular emotion recognition scenarios. 

\begin{figure}[h]
\centering
\includegraphics[width=3.5in]{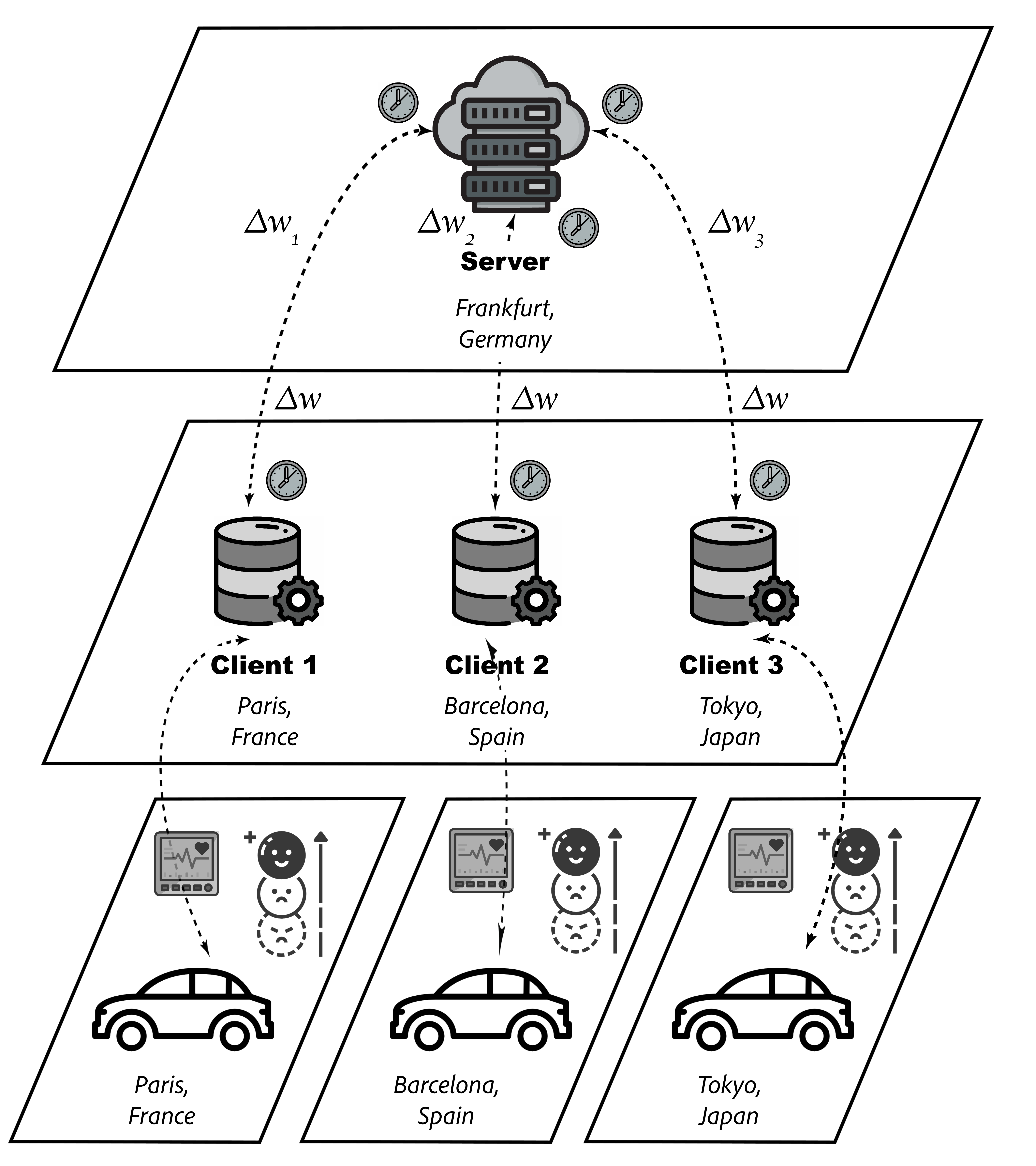}
\caption {\textit{Experimental setup}: Three geographically distributed clients extract physiological and emotion-related data from local vehicle cockpits. Local model updates are forwarded to the server for time-aware aggregation. A shared time reference is enforced across all nodes via explicit NTP-based synchronization, implemented using \textit{chrony}.}
\label{client2_comp}
\end{figure}

\subsection*{Use Case and System Architecture}

We consider a scenario of emotion recognition from physiological signals within vehicle cabins, a setting that demands high temporal sensitivity. Emotional states such as stress, fatigue, and excitement can shift rapidly in response to internal or external stimuli, making real-time recognition essential for both driver safety and user experience. To address this, our setup prioritizes model updates derived from the most recent data. We use the IAS Cockpit in-vehicle multimodal dataset, developed by the Institute of Industrial Automation and Software Engineering at the University of Stuttgart \cite{Guel24_2, Guel2024_1}, which contains physiological signals such as heart rate, skin conductance, and facial expression, key indicators of emotional state.

The experimental setup includes three geographically distributed clients, each representing a vehicle, communicating with a central server to form a synchronous federated learning architecture. Each client holds a subject-exclusive partition of the IAS Cockpit dataset: all recordings from a given driver are placed on a single client, resulting in non-overlapping data partitions with modest differences in size and class distribution, consistent with typical real-world heterogeneity. To reflect the communication characteristics of such a geographically distributed setup, latency is injected into the client-server communication layer using asyncio-based hooks, introducing controlled delays to emulate real-world network latency:

\begin{itemize}
    \item \textbf{Server location:} Frankfurt, Germany
    \item \textbf{Client 1:} Paris, France (ping $\approx$ 8.85 ms)
    \item \textbf{Client 2:} Barcelona, Spain (ping $\approx$ 23.349 ms)
    \item \textbf{Client 3:} Tokyo, Japan (ping $\approx$ 238.017 ms)
\end{itemize}

This setup simulates a realistic scenario where the clients experience varying network latencies, reflecting the challenges of edge-based learning in geographically distributed networks.

For emotion recognition, we designed a simple yet effective Multilayer Perceptron (MLP) using the TensorFlow Keras library. The model consists of three fully connected dense layers, tailored for a multi-class classification task. It is trained to predict one of six emotional states: happy, sad, angry, surprised, neutral, and fear. This architecture was selected to balance computational efficiency and accuracy, taking into account the constraints of FL in a dynamic, edge-computing environment.

\subsection*{Time Synchronization and Staleness Handling}

To maintain a consistent notion of time across the distributed clients and the server, we deploy \textit{Chrony}~\cite{chrony2025}, an implementation of NTP. The Chrony daemon, \textit{chronyd}, runs continuously on both clients and the server, synchronizing system clocks with external NTP servers and correcting for both offset and drift. In local-area network (LAN) deployments, a master–slave architecture may also be used, where one node (either client or server) acts as the time source for the others. However, given the distributed nature of our setup, each client independently synchronizes with publicly accessible NTP servers.

Table~\ref{tab:chrony_tokyo} presents a snapshot of \textit{Chrony’s} internal synchronization statistics for the Tokyo client, recorded at an arbitrary observation time. As the daemon continues to run over time, the update interval gradually increases, reflecting improved clock stability and reduced need for frequent adjustments.

Having established a shared time reference, each client timestamps its model update upon completion of local training. Upon receipt, the server computes the staleness of the update by comparing the client’s timestamp to its current system time. A corresponding freshness score is then derived using an exponential decay function, which down-weights stale updates. This score, in conjunction with the client’s dataset size, determines the aggregation weight during global model updates.

\begin{table}[h]
\centering
\caption{Chrony synchronization status for Tokyo client}
\begin{tabular}{l r}
\toprule
\textbf{Metric} & \textbf{Value} \\
\midrule
Stratum & 3 \\
System time offset & 0.00000039 seconds (fast) \\
Last offset & 0.000084246 seconds \\
RMS offset & 0.000084246 seconds \\
Frequency & 21.667 ppm slow \\
Residual frequency & +4.723 ppm \\
Skew & 0.410 ppm \\
Root delay & 0.000564836 seconds \\
Root dispersion & 0.027399288 seconds \\
Update interval & 2.0 seconds \\
Leap status & Normal \\
\bottomrule
\end{tabular}
\label{tab:chrony_tokyo}
\end{table}
\subsection*{Experimental Workflow}
The federated training proceeds in multiple synchronous communication rounds between clients and the central server, following these steps:

\begin{enumerate}
    \item Each client synchronizes its system clock using the Chrony protocol. If Chrony is unavailable, the local system time is used as a fallback.
    
    \item Clients collect and preprocess physiological data, then perform local model training on their respective private datasets.
    
    \item Upon completing local training, each client timestamps its model update and transmits both the update and timestamp to the central server.
    
    \item The server computes the \textit{staleness} of each update by comparing the received timestamp to its current system time.
    
    \item A corresponding \textit{freshness score} is computed using an exponential decay function applied to the staleness value.
    
    \item For each client, the server determines an aggregation weight based on both the freshness score and the size of the local dataset.
    
    \item The global model is updated by aggregating the received local models using the computed weights.
    
    \item The updated global model is then broadcast to all clients, initiating the next round of training.
    
    \item Steps 2--8 are repeated until the model converges or a predefined number of communication rounds is reached.
\end{enumerate}

We compare two distinct aggregation mechanisms:
\begin{table*}[t]
\centering
\caption{Comparison between SyncFed and FedAvg aggregation strategies}
\label{tab:syncfed_vs_fedavg}
\begin{tabular}{|l|p{6cm}|p{6cm}|}
\hline
\textbf{Aspect}                    & \textbf{SyncFed (Time-Aware Aggregation)}         & \textbf{FedAvg (Traditional Aggregation)}               \\[0.4ex]
\hline
\textbf{Model Aggregation}         & Recency- and size-weighted aggregation             & Size-proportional aggregation                           \\[0.4ex]
\hline
\textbf{Handling of Stale Clients} & Reduced influence of older or delayed updates      & Equal influence of all updates                          \\[0.4ex]
\hline
\textbf{Age of Information (AoI)}  & Lower AoI through prioritization of recent updates & AoI unaffected by update timeliness                      \\[0.4ex]
\hline
\textbf{Model Performance}         & Improved adaptability under variable data freshness & Performance driven by aggregate data volume             \\[0.4ex]
\hline
\textbf{Latency}                   & Fewer rounds under heterogeneous delays            & Increased rounds with asynchronous updates               \\[0.4ex]
\hline
\textbf{Scalability}               & Efficient scaling in large, distributed deployments & Overhead increase in large or distributed settings       \\[0.4ex]
\hline
\textbf{Robustness}                & Enhanced stability amid irregular client participation & Vulnerability to irregular or delayed contributions  \\[0.4ex]
\hline
\textbf{Convergence Speed}         & Faster convergence with time-aware prioritization  & Convergence rate tied to uniform, timely updates         \\[0.4ex]
\hline
\textbf{Implementation Complexity} & Moderate complexity from decay-based weighting     & Minimal complexity with straightforward averaging        \\[0.4ex]
\hline
\textbf{Recommended Context}       & Latency-sensitive, dynamic environments            & Homogeneous, synchronous settings                        \\[0.4ex]
\hline
\end{tabular}
\end{table*}

\begin{itemize}
    \item \textbf{SyncFed (Proposed):} Model updates are weighted using a hybrid metric combining the local dataset size and the freshness score. This ensures that newer updates are prioritized, leading to improved global model relevance and accuracy.
    \item \textbf{FedAvg (Baseline):} A traditional federated learning approach where updates are weighted solely based on the local dataset size, treating all updates equally regardless of freshness.
\end{itemize}

This comparison is also extended to multiple aspects as shown in Table~\ref{tab:syncfed_vs_fedavg}, verifying our results.

\subsection*{Results}

To compare SyncFed and FedAvg, we monitor two key metrics throughout training: model accuracy and Age of Information (AoI). Model accuracy is evaluated on a held-out validation set after each communication round, while AoI quantifies the time elapsed since a local update was generated—lower AoI values indicate fresher, more relevant data. These metrics enable a meaningful comparison of performance, particularly in assessing how each method handles stale updates in time-sensitive federated learning scenarios.

Training was conducted over 20 rounds across three clients. The results demonstrate a consistent advantage for SyncFed in maintaining higher accuracy over time and lower AoI values. As shown in Fig.~\ref{client2_comp}, FedAvg exhibited gradual and occasionally inconsistent accuracy improvements across rounds. Accuracy gains were slow, with noticeable plateaus—particularly in later rounds—due to FedAvg’s equal weighting of updates, regardless of freshness. This approach fails to account for stale updates arising from client-side latency or training delays, leading to slower convergence and reduced overall accuracy.

In contrast, SyncFed prioritizes recent updates by incorporating a freshness-aware weighting scheme. This strategy yields more effective model refinement, with clients achieving higher accuracy earlier and converging more reliably toward an optimal accuracy of 66\% by the final rounds. For comparison, benchmark studies on in-vehicle emotion recognition using skin conductance alone report accuracies between 33.1\% and 35.8\% \cite{Emotion_Benchmark}, highlighting the benefit of SyncFed’s update relevance filtering in federated contexts.

\begin{figure}[!h]
\centering
\includegraphics[width=3.5in,height=2.6in,keepaspectratio]{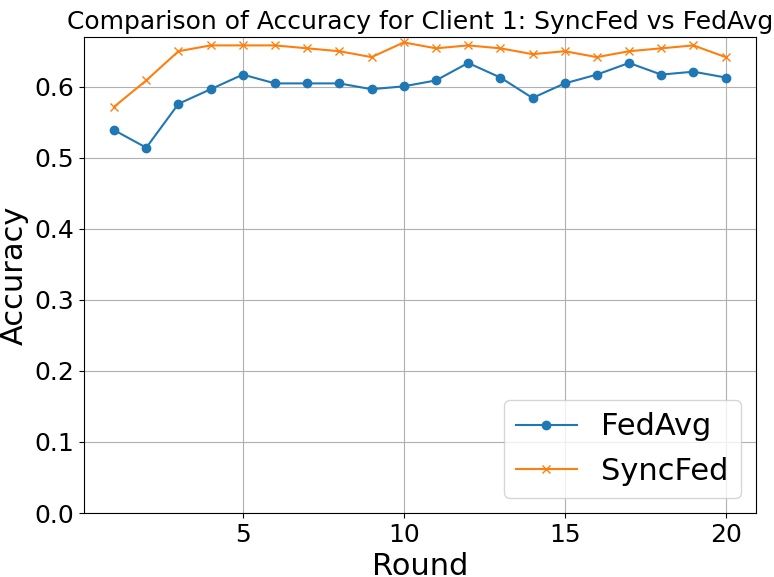}
\caption{Accuracy comparison of FedAvg and SyncFed for Client 1 across training rounds.}
\label{client2_comp}
\end{figure}

\begin{figure}[!h]
\centering
\includegraphics[width=3.5in,height=2.6in,keepaspectratio]{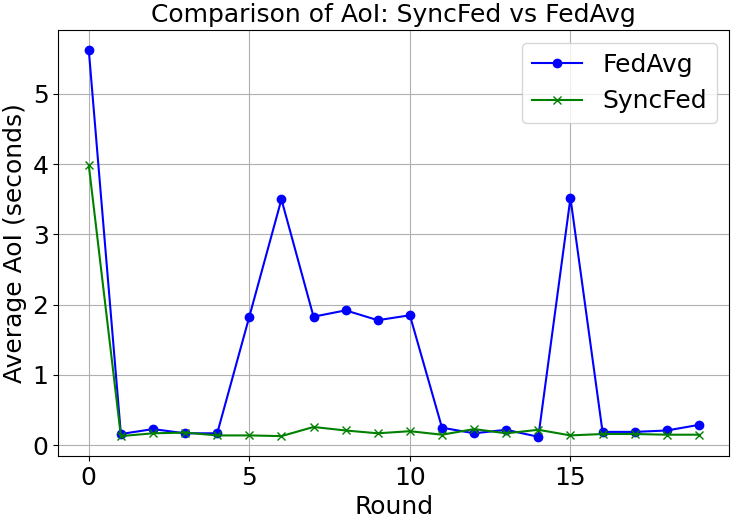}
\caption{Age of Information (AoI) Comparison between SyncFed and FedAvg across Rounds.}
\label{AoI_comparison}
\end{figure}


In addition to accuracy, we tracked AoI to quantify the freshness of updates incorporated during aggregation. As shown in Fig.~\ref{AoI_comparison}, SyncFed consistently achieves lower AoI values compared to FedAvg, reflecting its emphasis on prioritizing recent updates. This results in faster convergence and improved generalization. In contrast, FedAvg aggregates updates without regard to staleness, making it susceptible to performance degradation in the presence of heterogeneous client latencies. SyncFed’s time-aware aggregation mitigates this issue by down-weighting outdated contributions, ensuring that the global model reflects more current and relevant information.

Our results demonstrate that SyncFed improves global model performance in heterogeneous network environments by mitigating the influence of stale updates. Unlike FedAvg, SyncFed assigns reduced weights to outdated contributions, which proves especially beneficial in edge networks characterized by variable latencies. By reducing AoI and enabling freshness-aware aggregation, SyncFed stabilizes training dynamics and enhances model accuracy. Notably, these gains are achieved without introducing additional communication or computational overhead, making SyncFed well-suited for resource-constrained, latency-sensitive applications such as IoT and mobile federated learning systems.

\section{Discussion}
\label{discussion}

Additional aspects of time synchronization and deployment warrant further attention in the broader context of distributed learning. Complementing the design and evaluation of our approach, this section outlines considerations and possible directions for subsequent work.

\subsection{Accuracy of Synchronization and Logical Time}

Although NTP is frequently employed for its simplicity and low resource overhead, its synchronization accuracy is inherently limited by factors such as asymmetric delays, network jitter, and clock drift across participating devices. These limitations introduce discrepancies in physical timestamps that may, in turn, affect update ordering and model convergence. In practice, no synchronization mechanism is perfect, and devices can never be entirely in sync. In this work, we assume that synchronization accuracy always exceeds the system's minimum time resolution—that is, no event, message, or update occurs faster than the synchronization margin. If this assumption is violated, such events would be treated as concurrent and would necessitate the aggregation entity to invoke context-aware resolution mechanisms. In distributed computing, the limitations of physical time gave rise to logical time mechanisms such as Lamport timestamps~\cite{lamport} and vector clocks~\cite{vector}; unlike physical time, which targets real-world chronology, logical time captures causal relationships between events, supporting event ordering without a shared time base. This concept parallels the round-based semantics of FL, which implicitly absorb staleness through repeated iterations.

\subsection{Resource Constraints and Practical Deployment}

Our framework assumes that all participating clients are NTP-capable. However, in realistic deployment, particularly those involving resource-constrained IoT devices, this assumption may not hold. Many such devices lack the hardware necessary to support protocol-level synchronization. In these cases, synchronization can be delegated to external hubs or edge gateways. A similar architecture has been employed in distributed automotive testing, where gateways co-located with electronic control units (ECUs) functioned as synchronization hubs within a globally integrated testing stream ~\cite{ST1, ST2}. Prospectively, enhanced hardware capabilities and hardware-assisted timestamping could enable the use of more accurate, albeit less distributed-friendly, synchronization protocols such as Time-Sensitive Networking (TSN).

\subsection{Temporal Accuracy Versus System Overhead}

Incorporating explicit synchronization and time-awareness into FL frameworks introduces trade-offs between improved temporal accuracy and increased processing overhead. Frequent exchange and interpretation of timestamp data may increase latency and resource usage, potentially offsetting synchronization gains. This dynamic reflects a classical trade-off seen in distributed systems: the \textit{\textbf{probe effect}}~\cite{probe}, where the very act of monitoring or synchronizing a system can interfere with its performance. Nevertheless, we argue that the benefits of quantifiable staleness and principled temporal reasoning justify the overhead cost. Although FL is inherently more tolerant of delays than hard real-time systems, it still stands to benefit from synchronization mechanisms that enhance the consistency and reliability of model aggregation.

\section{Conclusion and Future Work}
\label{conclusion}
As Federated Learning (FL) continues to scale across heterogeneous and geographically dispersed environments, temporal coordination is becoming critical for model training. Current FL approaches rely on coarse, round-based heuristics to reason about staleness and lack mechanisms for time alignment or freshness quantification. In light of this limitation, we propose \textit{\textbf{SyncFed}}, a time-aware framework for distributed learning that integrates the following components:

\begin{itemize}
    \item \textit{\textbf{Explicit physical timestamping via NTP}}, as opposed to implicit versioning or round counters that fail to reflect temporal misalignment.
    \item \textit{\textbf{A unified temporal reference across all clients}}, as opposed to local clocks or asynchronous arrival times without system-wide coordination.
    \item \textit{\textbf{Quantifiable staleness and temporally informed update weighting}}, as opposed to binary inclusion criteria that overlook update freshness.
\end{itemize}
    
Our evaluation among three geographically distributed clients over 20 training rounds indicates that \textbf{SyncFed} performs favorably compared to the FedAvg baseline. It achieves faster convergence and maintains higher accuracy throughout training, showing improvements over benchmarks based solely on skin conductance for in-vehicle emotion recognition. Moreover, lower \textit{Age of Information} (AoI) values suggest that \textbf{SyncFed} makes more effective use of timely updates.
Future work will explore the integration of privacy-preserving mechanisms such as differential privacy, as well as using hardware-based timestamping to further enhance the accuracy of synchronization in aggregation.

\bibliographystyle{unsrtnat}
\bibliography{references}

\end{document}